\documentclass[lettersize,journal]{IEEEtran}
\usepackage{amsmath}
\usepackage[pdftex]{graphicx}
\usepackage{algorithm} 
\usepackage{booktabs}
\usepackage{mathtools}
\usepackage{array}
\usepackage{multirow}
\usepackage{multicol}
\usepackage[caption=false,font=normalsize,labelfont=sf,textfont=sf]{subfig}
\usepackage{url}
\usepackage{amsthm}

\usepackage[noabbrev]{cleveref}
\usepackage{amsfonts}

\usepackage{subfig}
\usepackage[autostyle]{csquotes}  
\usepackage{algpseudocode} 
\usepackage{mwe}
\usepackage{physics}
\usepackage{xcolor}

\usepackage{enumitem}
\setlist[enumerate]{leftmargin=*}
\setlist[itemize]{leftmargin=*}

\usepackage{xcolor}

\begin{document}

\title{From Wide to Deep: Dimension Lifting Network for Parameter-efficient Knowledge Graph Embedding}

\author{Borui~Cai,
        Yong~Xiang,~\IEEEmembership{Senior Member,~IEEE},
        Longxiang~Gao,~\IEEEmembership{Senior Member,~IEEE},
        Di~Wu,~\IEEEmembership{Member,~IEEE}
        He~Zhang,
        Jiong~Jin,~\IEEEmembership{Member,~IEEE}
        and~Tom~Luan,~\IEEEmembership{Senior Member,~IEEE}

\thanks{This work has been submitted to the IEEE for possible publication. Copyright may be transferred without notice, after which this version may no longer be accessible.} 
     
\thanks{B. Cai and Y. Xiang are with the School of Information Technology, Deakin University, VIC 3125, Australia.
E-mail: \{b.cai,yong.xiang\}@deakin.edu.au.}

\thanks{L. Gao is with Qilu University of Technology (Shandong Academy of Sciences), China. E-mail: gaolx@sdas.org.}

\thanks{D. Wu is with School of Mathematics, Physics and Computing, University of Southern Queensland, QLD 4350, Australia. E-mail: di.wu@unisq.edu.au.}

\thanks{H. Zhang is with CNPIEC KEXIN LTD, China. E-mail:  zhanghe@kxsz.net.}

\thanks{J. Jin is with the School of Science, Computing and Engineering Technologies, Swinburne University of Technology, VIC 3122, Australia. E-mail: jiongjin@swin.edu.au.}

\thanks{T. Luan is with the School of Cyber Engineering,
Xidian University, Xi’an, 710071, China. E-mail: tom.luan@xidian.edu.cn.}

}

\markboth{Journal of \LaTeX\ Class Files,~Vol.~14, No.~8, August~2021}%
{Shell \MakeLowercase{\textit{et al.}}: A Sample Article Using IEEEtran.cls for IEEE Journals}

\maketitle

\begin{abstract}
Knowledge graph embedding (KGE) that maps entities and relations into vector representations is essential for downstream applications. Conventional KGE methods require high-dimensional representations to learn the complex structure of knowledge graph, but lead to oversized model parameters. Recent advances reduce parameters by low-dimensional entity representations, while developing techniques (e.g., knowledge distillation or reinvented representation forms) to compensate for reduced dimension. However, such operations introduce complicated computations and model designs that may not benefit large knowledge graphs.
To seek a simple strategy to improve the parameter efficiency of conventional KGE models, we take inspiration from that deeper neural networks require exponentially fewer parameters to achieve expressiveness comparable to wider networks for compositional structures. We view all entity representations as a single-layer embedding network, and conventional KGE methods that adopt high-dimensional entity representations equal widening the embedding network to gain expressiveness. To achieve parameter efficiency, we instead propose a deeper embedding network for entity representations, i.e., a narrow entity embedding layer plus a multi-layer dimension lifting network (LiftNet). Experiments on three public datasets show that by integrating LiftNet, four conventional KGE methods with 16-dimensional representations achieve comparable link prediction accuracy as original models that adopt 512-dimensional representations, saving $68.4\%$ to $96.9\%$ parameters.
\end{abstract}



\section{Introduction}
\label{sect:1}
\IEEEPARstart{K}{nowledge} graphs containing a large number of facts benefit various downstream applications, ranging from open-domain question answering \cite{answer}, content-based recommender systems \cite{recommendation}, to text-centric information retrieval \cite{ir}. A fact in a knowledge graph is represented as a triple, which includes a head/subject entity, a tail/object entity, and the relation identifying the relationship between them; for example, $\{Da\ Vinci, painted, Mona\ Lisa\}$. Due to the complex multi-relational structure, directly applying knowledge graphs to downstream tasks is difficult. Knowledge Graph Embedding (KGE) methods \cite{transe} instead map entities and relations of a knowledge graph into vector representations, which preserve structural information of the knowledge graph and are convenient to be used by downstream tasks.

\begin{figure}[t!]
\centering
\subfloat[]{\includegraphics[width=1.62in]{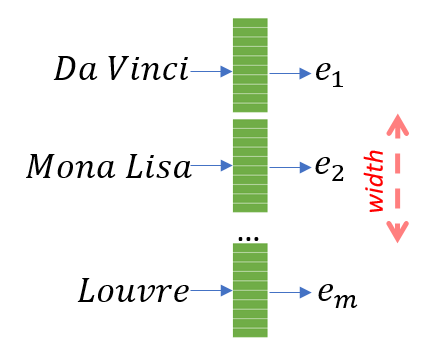}} \hspace{0.0in}
\subfloat[]{\includegraphics[width=1.82in]{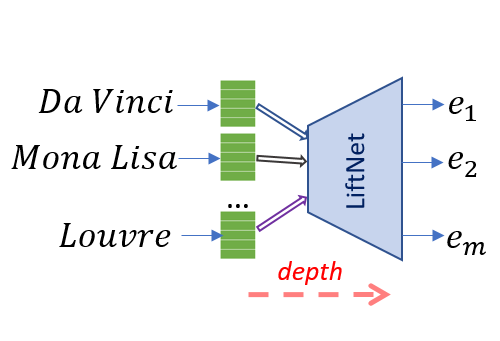}} \\
\caption{In (a), conventional KGE models that use high-dimensional entity representations equal to enlarging the width of the embedding layer. But we tend to achieve parameter efficiency by increasing the depth of the embedding network, i.e., a narrower embedding layer (low-dimensional entity representations) plus the LiftNet as shown in (b).}
\label{fig:intro}
\end{figure}

To accurately preserve the complex structural information of knowledge graphs, conventional KGE methods (e.g., TransE \cite{transe}) seek to increase the embedding dimension for better expressiveness and adopt high-dimensional entity/relation representations. However, since the scale of model parameters grows linearly with the representation dimension, this strategy leads to huge memory consumption, especially for large-scale real-life knowledge graphs. For example, when the embedding dimension is 512, a KGE model for Google knowledge graph \cite{google} (covering 5 billion entities by 2020) requires over $2,560$ billion model parameters; that is even 14 times larger than GPT-3 (one of the largest language models) \cite{gpt3}. This problem greatly limits KGE models on resource-constrained platforms. 

Some methods attempt to improve parameter efficiency by reducing the embedding dimension, and the key problem is how to maintain model expressiveness. Meanwhile, since entities normally are much more than relations in many knowledge graphs, they mostly reduce the dimension of entity representations. One type of method adopts knowledge distillation to transfer knowledge from pre-trained high-dimensional KGE models to low-dimensional entity representations \cite{dist2,dist3}. However, pre-trained high-dimensional KGE models are not always available, and such a training process also demands a huge memory footprint. Other methods reinvent more expressive representation forms, e.g., hyperbolic representation \cite{hyper}, or compositional anchor nodes \cite{anchor}; but these operations introduce complicated geometric computations and model designs that may not benefit large knowledge graphs.

In search of a simple way to reduce entity parameters and hence improve the parameter efficiency of conventional KGE models, we take inspiration from earlier discussions on wider vs deeper neural networks.
Study shows that deeper neural networks require exponentially fewer model parameters than wider networks to provide similar expressiveness for compositional structures \cite{wide1,deep1}, which widely exists in knowledge graphs \cite{code}. From this perspective, we view the concatenation of all entity representations as a single-layer embedding network. Then, a conventional KGE model that requires high-dimensional entity representations equals a rather wide network, i.e., by enlarging the width of the embedding layer to gain model expressiveness (Fig. \ref{fig:intro} (a)). Instead, we argue to simply achieve parameter efficiency by increasing the depth of the embedding network (Fig. \ref{fig:intro} (b)). 
Specifically, our deeper entity embedding network includes 1) an input layer that receives low-dimensional ($\hat{n}$) entity representations; 2) multiple hidden layers following the input layer to increase the model expressiveness; 3) an output layer that produces high-dimensional ($n$) entity representations, which can be directly adopted by existing KGE methods. 
We denote the entity embedding network as dimension-lifting network (LiftNet) since it lifts low-dimensional entity representations to high-dimensional. 
We refrain applying LiftNet to relation to accommodate conventional KGE models that design relation as other type of operations, e.g., translation on hyperplanes \cite{transh}.
The number of parameters for LiftNet-based entity representations becomes $|\mathcal{E}|\hat{n}+P_{net}$. $|\mathcal{E}|$ is the number of entities in the knowledge graph and $P_{net}$ is the model parameter size of LiftNet, which is negligible compared to $|\mathcal{E}|\hat{n}$ for large knowledge graphs. LiftNet can be conveniently integrated with conventional KGE models to make them parameter-efficient.
We evaluate the effectiveness by integrating LiftNet with four strong KGE methods and running experiments on three knowledge graph datasets of different sizes. The results show that by integrating with LiftNet, conventional KGE methods only require 16-dimensional entity representations to achieve link prediction accuracy comparable to original models of 512-dimensional, saving $68.4\%$ to $96.9\%$ model parameters.

\section{Related Work}

\subsection{Knowledge graph embedding}
KGE methods learn vector representations for knowledge graphs, and we roughly categorize them into three types. First, distance-based methods describe a fact with mathematical operations, e.g., TransE defines a relation as the translation. To better model N-N relations, TransH \cite{transh} and STransE \cite{stranse} project entities to relation-aware subspace with hyperplanes and matrices, respectively. Operations in the complex space \cite{rotate} or the polar coordinate system \cite{hake} are also introduced to improve flexibility. 
Second, tensor factorization methods, e.g., RESCAL \cite{rescal}, model the knowledge graph as a three-way tensor, and apply tensor decomposition to obtain entity/relation representations. Later methods further improve the effectiveness of decomposition. For example, DistMult \cite{distmult} applies an efficient diagonal relation matrix, ComplEx \cite{complex} introduces complex embedding to capture relation asymmetry, and SimplE \cite{simple} addresses the independence of entity embedding. 
Third, deep learning methods adopt deep neural networks to capture the complex relationships of entities and relations. ConvE \cite{conve}, and CapsE \cite{capsule} learn the complex interactions between entities and relations through convolutional layers and capsule layers, respectively. CompGCN \cite{compgcn} leverages GCN layers with entity-relation composition operations to capture interactions among entities and relations.

\subsection{Parameter-efficient knowledge graph embedding}
Since the scale of model parameters is linear with the embedding dimensions, many works focus on employing different techniques to improve the effectiveness of low-dimensional entity representations. 
Knowledge distillation is adopted to train low-dimensional representations with multiple pre-trained KGE models \cite{dist2}, which shows better performance than training from scratch. The number of pre-trained models is further reduced to one \cite{dist3} by introducing the dual influence between the pre-trained and the target models. Meanwhile, to accelerate the distillation, the low-dimensional model training is replaced by more efficient feature pruning of pre-trained models \cite{green}.
However, the obtained low-dimensional entity representations still show degraded performance.
To improve the performance, ROTH \cite{hyper} adopts more flexible hyperbolic space (compared to Euclidean space) for low-dimensional entity representations, but that also brings more complex hyperbolic geometry \cite{nonhyer}. To avoid that, contrastive learning is employed to flexibly control the strength of penalties for easy/difficult instances \cite{dist1}. Other works reinvent the entity representations as more expressive forms. For example, through learning a vocabulary of abstractive codewords that capture shared features of entities, entity representations become the composition of relevant codewords \cite{anchor,code}.

\section{The Proposed Method}

\subsection{Preliminaries}
A knowledge graph is a multi-relational graph denoted as $\mathcal{G} = (\mathcal{E}, \mathcal{R}, \mathcal{D})$. $\mathcal{E}$ and $\mathcal{R}$ are the collections of entities and relations, respectively, and $\mathcal{D}$ is the collection of facts contained in the knowledge graph. A fact $d\in \mathcal{D}$ is a triple denoted as $\{h,r,t\}$. In $d$, $h\in \mathcal{E}$ is the head entity, $t\in \mathcal{E}$ is the tail entity, and $r\in \mathcal{R}$ is the relation between the head entity and tail entity.
Knowledge graph embedding aims at learning vector representations for entities and relations while preserving the structural information of the knowledge graph. We denote the vector representations corresponding to $\{h,r,t\}$ as $\{e_{h},r_{r},e_{t}\}$. Conventional KGE methods map $h,r$, and $t$ into the same latent space, i.e., $e_{h},r_{r}$,$e_{t}\in \mathbb{R}^{n}$, and learn entity/relation representations by minimizing score functions defined as geometry operations; for example, $\l(h,r,t)=\|e_{h}+r_{r}-e_{t}\|_{p}$ (TransE).

Most KGE methods require relatively high embedding dimensions to sufficiently express the structural information of knowledge graph \cite{hyper}; however, that leads to oversized model parameters. 
Adopting low-dimensional ($\hat{e}\in \mathbb{R}^{\hat{n}}, \hat{n}\ll n$) entity representations can significantly reduce model parameters but results in low performance due to decreased expressiveness.
To achieve parameter-efficient KGE, we aim at using low-dimensional representations ($\hat{e}\in \mathbb{R}^{\hat{n}}, \hat{n}\ll n$), without degrading the performance. Following previous methods \cite{code,lre}, we focus on reducing the dimensions of entity representations since the number of entities is much larger than that of relations in many real-life knowledge graphs.

\subsection{Dimension Lifting Network}
As discussed above, we see the concatenation of entity representations as an embedding layer. To achieve parameter efficiency, we aim at replacing the wide embedding layer of high-dimensional entity representations ($\{e_{1}||e_{2}||...||e_{|\mathcal{E}|}\}$) with a deeper embedding network; that is, a narrow embedding layer of low-dimensional entity representations ($\{\hat{e}_{1}||\hat{e}_{2}||...||\hat{e}_{|\mathcal{E}|}\}$) and a multi-layer dimension lifting network (LiftNet).
$\{\hat{e}_{1}||\hat{e}_{2}||...||\hat{e}_{|\mathcal{E}|}\}$ and LiftNet are fully-connected; that means LiftNet can be interpreted as a dimension lifting function $f(*)$ that lifts $\hat{e}_{i}\in \mathbb{R}^{\hat{n}}$ to $e=f(\hat{e}_{i})\in \mathbb{R}^{n}$. $f(\hat{e}_{i})$ is used together with the original relation representations for score measurements and inference.

\begin{figure}[htbp]
\centering
\includegraphics[width=3.3in]{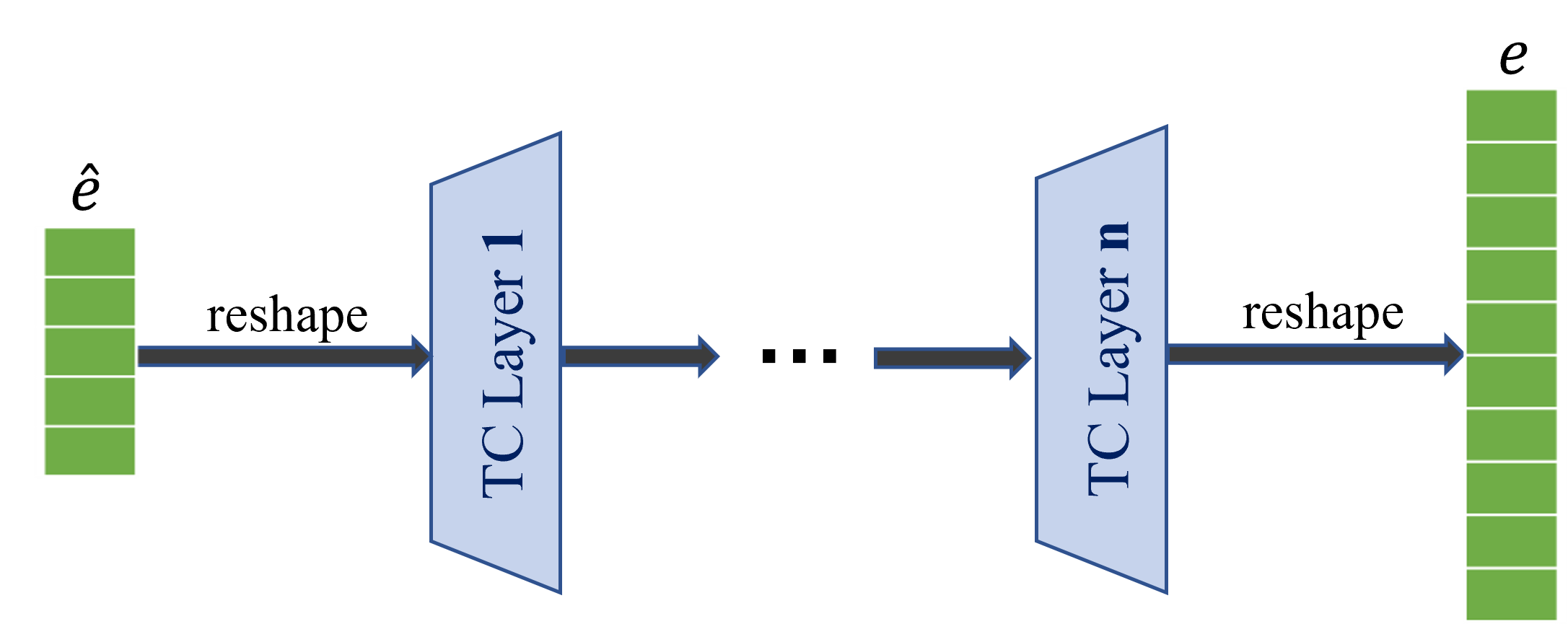} \\
\caption{The structure of LiftNet. $\hat{e}$ is the low-dimensional input entity representation, and LiftNet uses $n$ TC layers to progressively lift it to high-dimensional output $e$.}
\label{fig:frame}
\end{figure}

The main task is to design an effective $f(*)$ for KGE. 
An intuitive choice of $f(*)$ is multiple fully connected (FC) layers; however, FC layers require large numbers of parameters and are prone to overfitting for KGE \cite{conve}. Inspired by image processing, we refer to feature upsampling techniques for dimension lifting. 
Specifically, we adopt transposed convolution (TC) layer in LiftNet. A TC layer broadcasts the input elements via kernels, thus increasing the dimension of the output.
Different from traditional upsampling methods (e.g., nearest-neighbor, bilinear, and bicubic interpolation \cite{upsampling}), TC can capture the interactions among the parameters of entity representations. The structure of LiftNet is shown in Fig. \ref{fig:frame}. 

LiftNet adopts multiple TC layers to progressively lift the low-dimensional $\hat{e}$ to the high-dimensional $e$ (we find in experiments that LiftNet with only two FC layers already shows promising performance on three public knowledge graph datasets). 
LiftNet reshapes $\hat{e}$ as square feature maps in the forward pass and feeds them into the TC layers. After that, the representation is lifted into $\mathbb{R}^{c\times m_{1}\times m_{2}}$, where $c$ is the output channel number and $m_{1}\times m_{2}$ is the size of output feature map. The output of the last TC layer is then reshaped back to the expected high-dimensional $e$.
We use Tanh as the non-linear activation function between TC layers and also the output layer, considering that it is zero-centered and can capture complex feature patterns. We choose Adam  \cite{adam} as the optimizer for the model training. 

LiftNet is model-agnostic and can conveniently integrate with many conventional KGE methods for parameter-efficient representation learning, e.g., TransE, DistMult, by simply replacing the entity representation $e$ with $f(\hat{e})$. 
The score function of LiftNet-based method becomes $\l(h,r,t)=\l(f(\hat{e}_{h}),r_{r},f(\hat{e}_{t}))$, with $f(\hat{e}_{h}),r_{r},f(\hat{e}_{t})\in R^{n}$.
The parameter size of entity/relation representation for LiftNet-based methods is $|\mathcal{E}|\hat{n}+|\mathcal{R}|n+P_{net}$, where $P_{net}$ is the size of LiftNet parameters (mainly the transposed convolution kernels in TC layers) and $P_{net}\ll |\mathcal{E}|\hat{n}$. Therefore, for many real-world KG datasets that have far more entities than relations ($|\mathcal{E}|\gg |\mathcal{R}|$), LiftNet-based KGE methods only require approximately $\frac{\hat{n}}{n}$ parameters of original models that need high-dimensional entity representations.

\begin{table}[h]
\centering
\caption{Statistics of the datasets.}
\begin{tabular}{lccccc}
\toprule 
Dataset & \#E & \#R &Train/Valid/Test\\
\midrule
UMLS      &135  	&46   &5,327/569/633  \\ 
WN18RR    &40,943	&11   &86,835/3,034/3,134  \\
FB15K237  &14,541	&237  &272,115/17,535/20,466   \\ 
\bottomrule
\end{tabular}
\label{tab:dt}
\end{table}

\begin{table*}[h]
\centering
\caption{Link prediction accuracy of LiftNet methods and conventional KGE methods (the best in bold and the second-best underlined).}
\begin{tabular}{lccccccccc}
\toprule
\multirow{2}{*}{\textbf{Methods} (dim)} & \multicolumn{3}{c}{\textbf{UMLS}}& \multicolumn{3}{c}{\textbf{WN18RR}} & \multicolumn{3}{c}{\textbf{FB15K237}} \\
\cmidrule(lr){2-4}  \cmidrule(lr){5-7} \cmidrule(lr){8-10}
&MRR &H@1 &H@10 &MRR &H@1 &H@10 &MRR &H@1 &H@10 \\
\midrule
TransE (16)            &.568	&.372	&.844	&.102	&.020   &.281	&.138   &.074	&.261 \\
TransE (512)           &\underline{.661}	&\underline{.423}	&\textbf{.986}   &\textbf{.172}	&\textbf{.097}	&\underline{.392}   &\textbf{.208}	&\underline{.076}	&\textbf{.445} \\
LN-TransE (16)         &\textbf{.782}	&\textbf{.631}	&\underline{.970}   &\underline{.164}	&\underline{.082}	&\textbf{.406}   &\underline{.205}	&\textbf{.094}	&\underline{.412} \\
\midrule
TransH (16)            &.668	&.379	&.897   &.100	&.014	&.275   &.146	&.072	&.261 \\
TransH (512)           &\underline{.674}	&\underline{.513}	&\underline{.988}   &\underline{.167}	&\underline{.030}	&\underline{.393}   &\underline{.208}	&\underline{.084}	&\textbf{.445} \\
LN-TransH (16)         &\textbf{.848}	&\textbf{.730}	&\textbf{.989}   &\textbf{.174}	&\textbf{.032}	&\textbf{.417}   &\textbf{.235}	&\textbf{.134}	&\underline{.429} \\
\midrule
DistMult  (16)         &.100	&.012	&.270   &.015	&.004	&.034   &.031	&.011	&.057 \\
DistMult  (512)        &\underline{.655}	&\underline{.558}	&\underline{.846}   &\textbf{.388}	&\textbf{.329}	&\textbf{.477}   &\underline{.234}	&\underline{.146}	&\textbf{.407} \\
LN-DistMult  (16)      &\textbf{.710}	&\textbf{.614}	&\textbf{.891}   &\underline{.246}	&\underline{.296}	&\underline{.427}   &\textbf{.245}	&\textbf{.166}	&\underline{.405} \\
\midrule
ComplEx (16)           &.751	&.608	&.968   &.112	&.047	&.235   &.094	&.026	&.237 \\
ComplEx (512)          &\textbf{.870}	&\textbf{.781}	&\underline{.977}   &\textbf{.409}	&\textbf{.369}	&\textbf{.471}   &\underline{.259}	&\underline{.171}	&\textbf{.437} \\
LN-ComplEx (16)        &\underline{.806}	&\underline{.684}	&\textbf{.981}   &\underline{.332}	&\underline{.262}	&\underline{.441}   &\textbf{.266}	&\textbf{.178}	&\underline{.432} \\
\bottomrule
\end{tabular}
\label{tab:acc}
\end{table*}

\section{Experiments}

\subsection{Experiment Setup}
\paragraph{Datasets}
We adopt three public knowledge graph datasets widely used for KGE and link prediction, i.e., UMLS \cite{umls}, WN18RR \cite{transe}, and FB15K237 \cite{fb15k}. Detailed statistics of these datasets are summarized in Table \ref{tab:dt}. UMLS is a small dataset derived from the repository of biomedical vocabularies. WN18RR is a subset of WordNet, which describes hyponym and hypernym relations among words, and it removes inverse relations to avoid test leakage. FB15K237 adopts a similar process to remove nearly identical and inverse relations in FB15K \cite{transe}, which contains entity-relation triples of Freebase.

\begin{table}[h]
\centering
\caption{Link prediction accuracy of parameter-efficient KGE methods (the best in bold and the second-best underlined).}
\begin{tabular}{lcccccc}
\toprule
\multirow{2}{*}{\textbf{Methods}} & \multicolumn{2}{c}{\textbf{UMLS}}& \multicolumn{2}{c}{\textbf{WN18RR}} & \multicolumn{2}{c}{\textbf{FB15K237}} \\
\cmidrule(lr){2-3}  \cmidrule(lr){4-5} \cmidrule(lr){6-7}
&MRR &H@10 &MRR &H@10 &MRR &H@10 \\
\midrule
DualDE           &.763	&.967	&.168  &.389	&.157	&.374   \\
LightKG          &\textbf{.822}	&\underline{.970}	&.175   &.401	&.233	&.418   \\
GreenKGC         &.727	&.959	&\underline{.250}   &\underline{.429}	&\underline{.252}	&\textbf{.463}   \\
LN-ComplEx       &\underline{.806}	&\textbf{.981}	&\textbf{.332}   &\textbf{.441}	&\textbf{.266}	&\underline{.432}   \\
\bottomrule
\end{tabular}
\label{tab:acc1}
\end{table}

\paragraph{Evaluation metrics}
We evaluate the performance of KGE with link prediction \cite{transe}, which is a popular application of KGE. Specifically, the task is to predict missing entity given a query ($\{h,r,?\}$ or $\{?,r,t\}$). 
Following existing works \cite{distmult,complex}, we use mean reciprocal ranking (MRR) and H@k to measure the prediction accuracy. MRR is the average inverse rank for all test triples, and H@k is the percentage of ranks lower than or equal to k. The maximum values of MRR and H@k are both 1, and the higher MRR or H@k, the better the performance. We adopt the filtered setting \cite{hyper} to exclude candidate triples that have been seen in training, validation, and testing sets.

\paragraph{Conventional KGE models}
We choose a set of strong conventional KGE models to show the effectiveness of the proposed LiftNet method. That includes TransE, TransH, DistMult, and ComplEx. TransE and TransH are translational models that adopt distance measurements for related entities and their relations. DistMult and ComplEx aim at semantic matching and adopt tensor decomposition in real and complex spaces, respectively. Their scoring functions are summarized in Table \ref{tab:score}.
By replacing $e$ with $f(\hat{e})$ in the above four methods, we implement LN-TransE, LN-TransH, LN-DistMult, and LN-ComplEx for parameter-efficient KGE.

\begin{table}[h]
\centering
\caption{Scoring functions of experimented KGE methods.}
\begin{tabular}{lccccc}
\toprule 
Methods & Scoring Function $l(h,r,t)$\\
\midrule
TransE      &$\lVert e_{h}+r_{r}-e_{t}\rVert_{p}$  \\ 
TransH      &$\lVert e_{h,\perp}+r_{r}-e_{t,\perp}\rVert_{p},e_{\perp}=e-w_{r}ew_{r}^{T}$  \\
DistMult    &$\langle e_{h}, r_{r}, e_{t}\rangle$   \\ 
ComplEx     &$\langle e_{h}, r_{r}, e_{t}\rangle, r_{r}\in \mathcal{C}^{n}$ \\
\bottomrule
\end{tabular}
\label{tab:score}
\end{table}

\paragraph{Implementation details}
We implement all the methods with OpenKE \cite{openke}, which is a pytorch-based open-source framework for knowledge embedding\footnote{Codes are available at https://github.com/brcai/LiftNet}. We run TransE, TransH, DistMult, and ComplEx with low-dimensional (16) and high-dimensional (512) embedding dimensions to show the difference in link prediction performance. Meanwhile, we implement a two-layer LiftNet for LiftNet-based methods, which lifts the dimension of entity representations from 16 to 512. In LiftNet, the parameters of the two TC layers as set as \{InChan:1, OutChan:4, Kernel:3\} and \{InChan:4, OutChan:8, Kernel:3\}, respectively, with stride=1 and padding=0. We fix the random seed for all experiments and use MRR of the validation set to find the optimal learning rate from \{0.01, 0.05, 0.1, 0.5\}. The maximum epoch is 500.

\subsection{Main Results}
\paragraph{Link prediction accuracy}
The accuracy of link prediction of the proposed LifeNet-based methods and compared conventional methods are shown in Table \ref{tab:acc}. We observe that TransE, TransH, DistMult, and ComplEx all obtain higher link prediction accuracy with 512 embedding dimensions than with 16 embedding dimensions, due to the increased expressiveness of entity/relation representations. Especially on WN18RR and FB15K237 datasets, the accuracy of 512-dimensional models is far better than 16-dimensional models.
Meanwhile, LN-TransE, LN-TransH, LN-DistMult, and LN-ComplEx, which also adopt 16-dimensional entity representations, achieve more accurate prediction results than corresponding methods that adopt 16-dimensional representations. In fact, their results are comparable to original models with 512-dimensional representations. We observe that on UMLS and FB15K237 datasets, MRR, H@1, and H@10 obtained by LN-TransH are even higher than those of original TransH with 512 embedding dimension; LN-DistMult also outperforms DistMult with 512 embedding dimension on UMLS dataset. For the rest results, LiftNet-based methods only show link prediction accuracy that are slightly lower than original methods with 512 embedding dimension. 
We also observe that LiftNet methods perform better on UMLS and FB15K237 datasets than on WN18RR, because UMLS and FB15K237 contain richer structural information, i.e., the average edges per entity (39.5 and 18.7) are much higher than that of WN18RR (2.1).

We further compare the performance of LiftNet with existing parameter-efficient KGE methods on the three datasets. We select three strong parameter-efficient KGE methods of different types, i.e., DualDE \cite{dist2}, LightKG \cite{lightkg}, and GreenKGC \cite{green}, which performs distillation, quantization, and pruning, respectively. For a fair comparison, we adopt the same random negative sampling \cite{negative} and unify the base model of all parameter-efficient methods as ComplEx (since ComplEx performs better than TransE, TransH, and DistMult on the three datasets). We follow the model setup of LightKE \cite{lightkg} and adopt 16-dimensional entity representations for the rest parameter-efficient models. As shown in Table \ref{tab:acc1}, LN-ComplEx generally achieves more accurate link prediction than DualDE, LightKG, and GreenKGC. LN-ComplEx achieves the best link prediction accuracy on WN18RR dataset (both MRR and H@10), and the best or the second best on UMLS and FB15K237 datasets. In particular, LN-ComplEx outperforms DualDE on all datasets, and is slight lower than LightKG and GreenKGC on UMLS (MRR is smaller for 0.016) and FB15K237 (H@10 is smaller for 0.031), respectively.
\begin{table}[h]
\centering
\caption{Parameter efficiency (M is short for million).}
\begin{tabular}{lccc}
\toprule 
&UMLS & WN18RR & FB15K237 \\
\midrule
TransE (512)            &0.092M	&20.968M  &7.566M  \\
LN-TransE (16)	        &0.026M	&0.661M   &0.354M   \\
\scriptsize{\textit{Percentage}}        &\scriptsize{\textit{28.2\%}}	&\scriptsize{\textit{3.1\%}}  &\scriptsize{\textit{4.7\%}}  \\ 
\midrule
TransH (512)            &0.116M	&20.974M  &7.687M  \\
LN-TransH (16)	        &0.049M	&0.667M   &0.475M   \\ 
\scriptsize{\textit{Percentage}}        &\scriptsize{\textit{42.2\%}}	&\scriptsize{\textit{3.2\%}}  &\scriptsize{\textit{6.2\%}}  \\ 
\midrule
DistMult (512)            &0.092M	&20.968M  &7.566M  \\
LN-DistMult	(16)          &0.026M	&0.661M   &0.354M   \\ 
\scriptsize{\textit{Percentage}}        &\scriptsize{\textit{28.2\%}}	&\scriptsize{\textit{3.1\%}}  &\scriptsize{\textit{4.7\%}}  \\ 
\midrule
ComplEx (512)            &0.185M	&41.926M  &15.132M  \\
LN-ComplEx (16)	         &0.051M	&1.321M   &0.708M   \\ 
\scriptsize{\textit{Percentage}}        &\scriptsize{\textit{27.5\%}}	&\scriptsize{\textit{3.1\%}}  &\scriptsize{\textit{4.7\%}}  \\ 
\bottomrule
\end{tabular}
\label{tab:param}
\end{table}

\begin{table}[h]
\centering
\caption{Memory usage (MB is short for Megabyte).}
\begin{tabular}{lccc}
\toprule 
&UMLS & WN18RR & FB15K237 \\
\midrule
TransE (512)            &0.354MB	&80.022MB  &28.864MB  \\
LN-TransE (16)	        &0.101MB	&2.523MB    &1.353MB   \\
\scriptsize{\textit{Percentage}}        &\scriptsize{\textit{28.5\%}}	&\scriptsize{\textit{3.1\%}}  &\scriptsize{\textit{4.7\%}}  \\ 
\midrule
TransH (512)            &0.445MB	&80.043MB  &29.327MB  \\
LN-TransH (16)	        &0.191MB	&2.544MB   &1.816MB   \\ 
\scriptsize{\textit{Percentage}}        &\scriptsize{\textit{42.9\%}}	&\scriptsize{\textit{3.2\%}}  &\scriptsize{\textit{6.2\%}}  \\ 
\midrule
DistMult (512)            &0.354MB	&80.022MB  &28.864MB  \\
LN-DistMult	(16)          &0.101MB	&2.523MB   &1.353MB   \\ 
\scriptsize{\textit{Percentage}}        &\scriptsize{\textit{28.5\%}}	&\scriptsize{\textit{3.1\%}}  &\scriptsize{\textit{4.7\%}}  \\ 
\midrule
ComplEx (512)            &0.708MB	&160.043MB  &57.728MB  \\
LN-ComplEx (16)	         &0.199MB	&5.043MB   &2.704MB   \\ 
\scriptsize{\textit{Percentage}}        &\scriptsize{\textit{28.1\%}}	&\scriptsize{\textit{3.1\%}}  &\scriptsize{\textit{4.7\%}}  \\ 
\bottomrule
\end{tabular}
\label{tab:memory}
\end{table}

\paragraph{Parameter efficiency}
In Table \ref{tab:param}, we show the parameter efficiency of LiftNet-based methods on the three datasets. The results are shown pair-wisely, i.e., the numbers of parameters required by 512-dimensional KGE models and those of corresponding LiftNet models, because they achieve similar link prediction accuracy. KGE methods with 16-dimensional representations are excluded due to their unsatisfying link prediction accuracy.
As the results show, the LiftNet-based methods save $68.4\%$ to $96.9\%$ model parameters or the original models. Since the parameter size of LiftNet is negligible (324 parameters), the level of parameter efficiency is mainly affected by the ratio of entities to relations in the dataset. On the smallest UMLS dataset, which has the smallest entity/relation ratio (around $3$), LiftNet-based methods require approximately $28\%\sim 42\%$ model parameters of original KGE models. However, on larger WN18RR and FB15K237 datasets that have much larger entity/relation ratios (around $3722$ and $61$, respectively), they only require approximately $3.1\%$ and $5.1\%$ model parameters of the original KGE methods to achieve comparable link prediction accuracy. This makes LiftNet promising for real-world knowledge graphs such as Google knowledge graph that has billions of entities. 
We compare the memory usage of 512-dimensional KGE models and the corresponding LiftNet models. The results in Table \ref{tab:memory} show that adopting LiftNet can significantly reduce the memory usage, e.g., from around 160 MB (512-dimensional ComplEx) to 5 MB (16-dimensional LN-ComplEx) on WN18RR dataset. Meanwhile, the percentage of memory reduction is consistent to that of model parameters (Table \ref{tab:param}), especially on larger WN18RR and FB15K237 datasets, proving the reduction of memory usage is due to the fewer model parameters of LiftNet.

We further discuss the impact of LiftNet on training complexity since it introduces additional neural layers to obtain entity representations. Specifically, we train each method for 100 epochs and then provide the average training time taken per epoch. As shown in Table \ref{tab:runtime}, by adopting LiftNet, the ratio of training time increases ranges from 1.08 to 1.33 times, with respect to 0.878 seconds to 0.814 seconds (LN-TransE to TransE, WN18RR dataset) and 0.052 seconds to 0.039 seconds (LN-DistMult to DistMult, UMLS dataset).
The above results show that training time increases is insignificant compared to the improved parameter efficiency.

\begin{table}[h]
\centering
\caption{Average training time for one epoch (seconds).}
\begin{tabular}{lccc}
\toprule 
&UMLS & WN18RR & FB15K237 \\
\midrule
TransE (512)            &0.032	&0.814  &1.971   \\
LN-TransE (16)	        &0.036	&0.878  &2.107   \\
\scriptsize{\textit{Ratio}}        &\scriptsize{\textit{1.13}}	&\scriptsize{\textit{1.08}}  &\scriptsize{\textit{1.07}}  \\ 
\midrule
TransH (512)            &0.064	&1.073  &3.368  \\
LN-TransH (16)	        &0.081	&1.289  &4.156   \\ 
\scriptsize{\textit{Ratio}}        &\scriptsize{\textit{1.27}}	&\scriptsize{\textit{1.20}}  &\scriptsize{\textit{1.23}}  \\ 
\midrule
DistMult (512)            &0.039	&0.783  &2.392  \\
LN-DistMult	(16)          &0.052	&0.962  &3.038   \\ 
\scriptsize{\textit{Ratio}}        &\scriptsize{\textit{1.33}}	&\scriptsize{\textit{1.23}}  &\scriptsize{\textit{1.27}}  \\ 
\midrule
ComplEx (512)            &0.078	&1.589  &4.424  \\
LN-ComplEx (16)	         &0.101	&1.859  &5.470   \\ 
\scriptsize{\textit{Ratio}}        &\scriptsize{\textit{1.29}}	&\scriptsize{\textit{1.17}}  &\scriptsize{\textit{1.24}}  \\ 
\bottomrule
\end{tabular}
\label{tab:runtime}
\end{table}

\subsection{Sensitivity Analysis}
\paragraph{Input dimensions}
The input dimensions highly affect the parameter efficiency of LiftNet-based methods; thus we analyze its influence on the model performance. We evaluate LN-TransE, LN-TransH, LN-DistMult, and LiftNet-ComplEx with four input dimensions $\{4,16,64,256\}$ on WN18RR dataset. To do that, we adjust the setups of TC layers (channel number, kernel size, and stride) accordingly to lift the dimensions of entity representations to 512. Due to the characteristics of TC layers, a higher input dimension requires a lower number of parameters (fewer kernels) in LiftNet.

\begin{figure}[t!]
\centering
\subfloat[]{\includegraphics[width=1.7in]{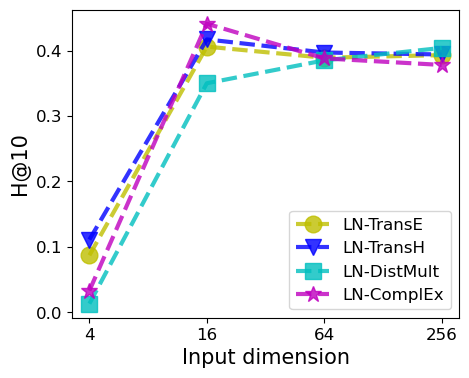}}
\subfloat[]{\includegraphics[width=1.7in]{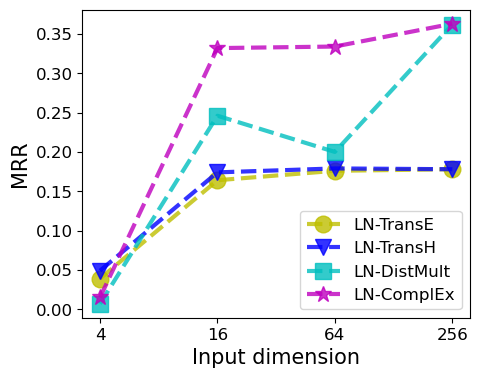}} \\
\caption{Link prediction accuracy w.r.t. input dimensions.}
\label{fig:in}
\end{figure}

The results of LiftNet-based methods for knowledge graph link prediction (accuracy measured by H@10 and MRR) are shown in Fig. \ref{fig:in}. Generally, on WN18RR datasets, we observe the link prediction accuracy increases with higher input dimension, and the increase is significant from 4-dimension to 16-dimension. However, after that, the accuracy only slightly increases or even drops, e.g., LN-ComplEx and LN-DistMult in Fig. \ref{fig:in} (a), with higher input dimensions partly due to over-parameterization. That shows the effectiveness of LiftNet with low-dimensional entity representations.

\begin{figure}[t!]
\centering
\subfloat[]{\includegraphics[width=1.7in]{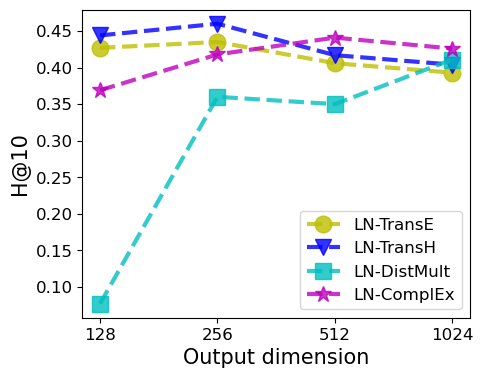}}
\subfloat[]{\includegraphics[width=1.7in]{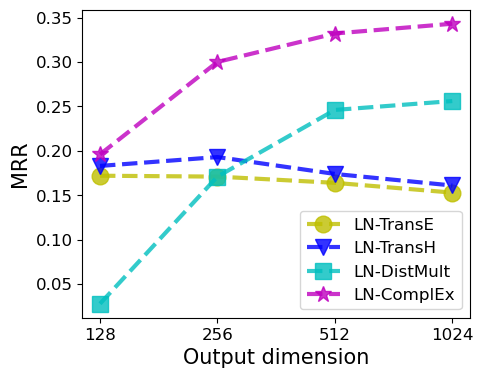}} \\
\caption{Link prediction accuracy w.r.t. output dimensions.}
\label{fig:out}
\end{figure}

\paragraph{Output dimension}
We show the sensitivity of the performance of LiftNet-based methods regarding the output dimensions on WN18RR dataset. To do that, we vary the output dimensions from 128 to 1024 and adjust the setups of the two TC layers in LiftNet accordingly. The results in Fig. \ref{fig:out} show two different trends. First, LN-TransE and LN-TransH show increased H@10 in Fig. \ref{fig:out} (a) and MRR in Fig. \ref{fig:out} (b) with higher output dimensions and then slightly decrease; while the results of LN-DistMult and LN-ComplEx mostly increase. The difference is caused by their different scoring functions that lead to different sensitivity to over-parameterization; addictive scoring functions for LN-TransE and LN-TransH and multiplicative scoring functions for LN-DistMult and LN-ComplEx. 
The above results generally support why LiftNet needs to "lift" output dimensions, rather than maintain or even reduce them.

\paragraph{Number of layers}
We further analyze the sensitivity of LiftNet regarding the number of TC layers on WN18RR dataset. For demonstration, we vary the number of TC layers in LiftNet-based methods from one to four. Note one layer LiftNet is still different from the linear transformation used in \cite{lre} as it has non-linear activation on the output layer.
We adjust the setups of TC layers by increasing/decreasing the kernel size accordingly to ensure the output dimensions are always 512. The results with respect to link prediction accuracy (H@10) are shown in Table \ref{tab:layer}. We see that the results of LN-TransE and LN-TransH are relatively stable, introducing more TC layers only slightly increases model performance. While LN-DistMult and LN-ComplEx are more sensitive to different layer numbers, their performance collapses with 4 TC layers partly due to training difficulties and overfitting.

\begin{table}[h]
\centering
\caption{The link prediction accuracy (by MRR) of LiftNet-based methods w.r.t. the number of TC layers.}
\begin{tabular}{ccccc}
\toprule 
Layers & LN-TransE & LN-TransH & LN-DistMult & LN-ComplEx\\
\midrule
1      &.395	&.412  &.088  &.474 \\
2	   &.406	&.417  &.427  &.441 \\ 
3      &.406	&.422  &.358  &.375 \\ 
4      &.411    &.421  &.004  &.010 \\
\bottomrule
\end{tabular}
\label{tab:layer}
\end{table}

\begin{table}[h]
\centering
\caption{The link prediction accuracy (by MRR) of LiftNet variants that adopt FC layers.}
\begin{tabular}{ccccc}
\toprule 
Layers & LN-TransE & LN-TransH & LN-DistMult & LN-ComplEx\\
\midrule
TC (2)      &.412	&.429  &.405  &.432 \\
\midrule
FC (2)	    &.352	&.402  &.408  &.411 \\ 
FC (3)      &.353	&.405  &.260  &.289 \\ 
FC (4)      &.355   &.404  &.124  &.180 \\
\bottomrule
\end{tabular}
\label{tab:var}
\end{table}

\subsection{LiftNet Variants}
In LiftNet, we adopt TC layers to progressively lift the dimensions. To demonstrate the effectiveness of such a design, we implement LiftNet variants with fully connected (FC) layers for comparison. The experiment is conducted on the largest FB15K237 dataset, with accuracy measured by MRR. Specifically, we include LiftNet variants of 2 to 4 FC layers, and the results are shown in Table \ref{tab:var}. We see that in most cases, LiftNet with 2 TC layers achieves more accurate link prediction results than LiftNet with 2 or more FC layers. We do not claim that TC layers are better than FC layers, but in KGE where the model is sensitive to insufficient expressiveness or over-parameterization, LiftNet with TC layers is relatively easier to achieve more accurate link prediction results.

\section{Conclusion}
In this paper, we propose a simple LiftNet method that helps improve the parameter efficiency of conventional KGE models. LiftNet adopts a multi-layer neural network to enhance the expressiveness of low-dimensional entity representations. Experiments conducted on three public knowledge graph datasets show that with LiftNet, conventional KGE models TransE, TransH, DistMult, and ComplEx can save up to $68.4\%$, $96.9\%$, and $94.9\%$ model parameters on UMLS, WN18RR, and FB15K237 datasets, respectively, in link prediction. 
One potential drawback of LiftNet is the increased training/inference time, due to the multi-layer back/forward propagation. However, experiments show that LiftNet requires only two TC layers for the tested datasets, hence the additional run-time complexity pales in comparison to the improved parameter efficiency. Meanwhile, LiftNet still requires input entity representations to generate desired output, in future work, we will explore semantic encoders that generate entity representations from their textual information (e.g., names and descriptions) for higher model effectiveness.

\ifCLASSOPTIONcaptionsoff
  \newpage
\fi

\bibliographystyle{IEEEtran}
\bibliography{ref.bib}

\end{document}